\documentclass{article}
\usepackage{arxiv}

\usepackage[utf8]{inputenc} 
\usepackage[T1]{fontenc}    
\usepackage{hyperref}       
\usepackage{url}            
\usepackage{booktabs}       
\usepackage{amsfonts}       
\usepackage{nicefrac}       
\usepackage{microtype}      
\usepackage{lipsum}		
\usepackage{graphicx}
\usepackage[numbers,sort&compress]{natbib}
\usepackage{doi}
\usepackage{CJKutf8}
\usepackage{pifont}
\usepackage{xcolor}
\usepackage{array}
\usepackage{multirow}

\title{Topic Modeling and Sentiment Analysis on Japanese Online Media's Coverage of Nuclear Energy}

\author{{Yifan Sun}\\
	Kyoto University\\
	Kumatori, Osaka 590-0494\\
	\texttt{sun.yifan.7r@kyoto-u.ac.jp} \\
	\And
	{Hirofumi Tsuruta} \\
	Sakura internet Inc.\\
	Osaka kita-ku, Osaka 530-0011 \\
	\texttt{hi-tsuruta@sakura.ad.jp} \\
	\And
	{Masaya Kumagai} \\
    Sakura internet Inc.\\
	Osaka kita-ku, Osaka 530-0011 \\
    $\&$ \\
	Kyoto University\\
	Kumatori, Osaka 590-0494\\
	\texttt{kumagai.masaya.3n@kyoto-u.ac.jp} \\
	\And
	{Ken Kurosaki} \\
	Kyoto University\\
	Kumatori, Osaka 590-0494\\
	\texttt{kurosaki.ken.6n@kyoto-u.ac.jp} \\
}

\date{}

\renewcommand{\shorttitle}{Nuclear Energy Topic Modeling and Sentiment Analysis}

\begin{document}
\maketitle

\begin{abstract}
	Thirteen years after the Fukushima Daiichi nuclear power plant accident, Japan's nuclear energy accounts for only approximately 6$\%$ of electricity production, as most nuclear plants remain shut down. To revitalize the nuclear industry and achieve sustainable development goals, effective communication with Japanese citizens, grounded in an accurate understanding of public sentiment, is of paramount importance. While nationwide surveys have traditionally been used to gauge public views, the rise of social media in recent years has provided a promising new avenue for understanding public sentiment. To explore domestic sentiment on nuclear energy-related issues expressed online, we analyzed the content and comments of over 3,000 YouTube videos covering topics related to nuclear energy. Topic modeling was used to extract the main topics from the videos, and sentiment analysis with large language models classified user sentiments towards each topic. Additionally, word co-occurrence network analysis was performed to examine the shift in online discussions during August and September 2023 regarding the release of treated water. Overall, our results provide valuable insights into the online discourse on nuclear energy and contribute to a more comprehensive understanding of public sentiment in Japan. 
\end{abstract}


\section{Introduction}
In the aftermath of the Fukushima Daiichi (1F) nuclear power plant accident, all nuclear reactors in Japan were temporarily shut down for inspections, leaving the country without nuclear energy in its electricity production for the first time since the 1970s \cite{ANRE2021}. Although nuclear power has gradually returned to Japan’s energy mix, accounting for 6$\%$ of electricity with 13 operational reactors as of November 2024, the Ministry of Economy, Trade, and Industry (METI) has set a target to increase this share to 20–22$\%$ by 2030 \cite{METI2022}. To further expand nuclear energy amid ongoing safety concerns following the Fukushima 1F accident, it is necessary for the government to build public trust, a responsibility recently formalized in an amendment to the Atomic Energy Basic Act.

To foster effective communication with the public, nationwide polls have long been a primary tool for gauging sentiment on nuclear energy topics. Major news outlets such as NHK \cite{nhk2022}, Asahi Shimbun \cite{asahi2024}, and Nikkei \cite{nikkei2023} have conducted surveys on specific issues, while institutions like the Institute of Nuclear Safety System \cite{kitada2016public} and the Japan Atomic Energy Relations Organization have carried out comprehensive annual surveys. However, traditional polls face inherent limitations, including restricted reach and ambiguous interpretations of nonsubstantive responses, highlighting the need for supplementary methods to improve understanding of public sentiment.

In recent years, the rise of social media platforms has accumulated an abundance of user comments and posts, offering a promising avenue for understanding public opinions on nuclear energy. Early studies, such as those by Ikegami et al. \cite{ikegami2013topic} and Kim $\&$ Kim \cite{kim2014public}, introduced the use of sentiment orientation dictionaries to analyze Tweets related to the Fukushima 1F accident, a method that has since gained widespread adoption. For example, Park \cite{park2019positive} and Jeong et al. \cite{jeong2021sentiment} employed sentiment dictionaries to examine online sentiment toward major nuclear-related issues in Korea, while Pu et al. \cite{pu2022chinese} applied a similar approach to gauge public opinion on the release of treated water on the Chinese social media platform Weibo. In Japan, Hasegawa et al. \cite{hasegawa2020changing} analyzed 19 million Japanese Tweets to assess public sentiments surrounding radiation and Fukushima one year after the accident.

While prior studies have provided valuable insights into online sentiment toward nuclear energy, based on data at a scale unattainable through traditional polls, two key issues limit their accuracy and impact. First, lexicon-based sentiment analysis often struggles with accuracy due to its reliance on static word sentiments \cite{kim2014public}, failing to account for context or detect nuanced emotions like irony and sarcasm. Second, identifying the specific nuclear energy issues in online posts remains challenging, especially due to their short-format nature. For instance, Park \cite{park2019positive} and Jeong et al. \cite{jeong2021sentiment} manually identified nuclear-related events for sentiment assignment, introducing potential human bias. Recent work by Kwon et al. \cite{kwon2024sentiment} attempted to address these challenges by incorporating Large Language Models (LLMs), which are better equipped to understand human language, to preform both sentiment analysis and categorize Tweets into themes such as political or energy-related. However, it still lacks the granularity needed to link sentiments to specific issues regarding nuclear energy.

To fully address these challenges in providing a granular understanding of online sentiment on nuclear energy, this work applies state-of-the-art LLMs for sentiment analysis while shifting the focus from analyzing user comments on Twitter to YouTube, a platform that has received comparatively little attention. Unlike Tweets, YouTube comments offer a unique advantage in that they are closely tied to videos content, directly linking user's sentiments and opinions to specific topics discussed. Consequently, this work seeks to answer the following question: What are the main topics presented in nuclear energy-related news videos on YouTube, and what are viewers' sentiments toward these topics?

\begin{figure}[h]
    \centering
    \includegraphics[width=0.6\linewidth]{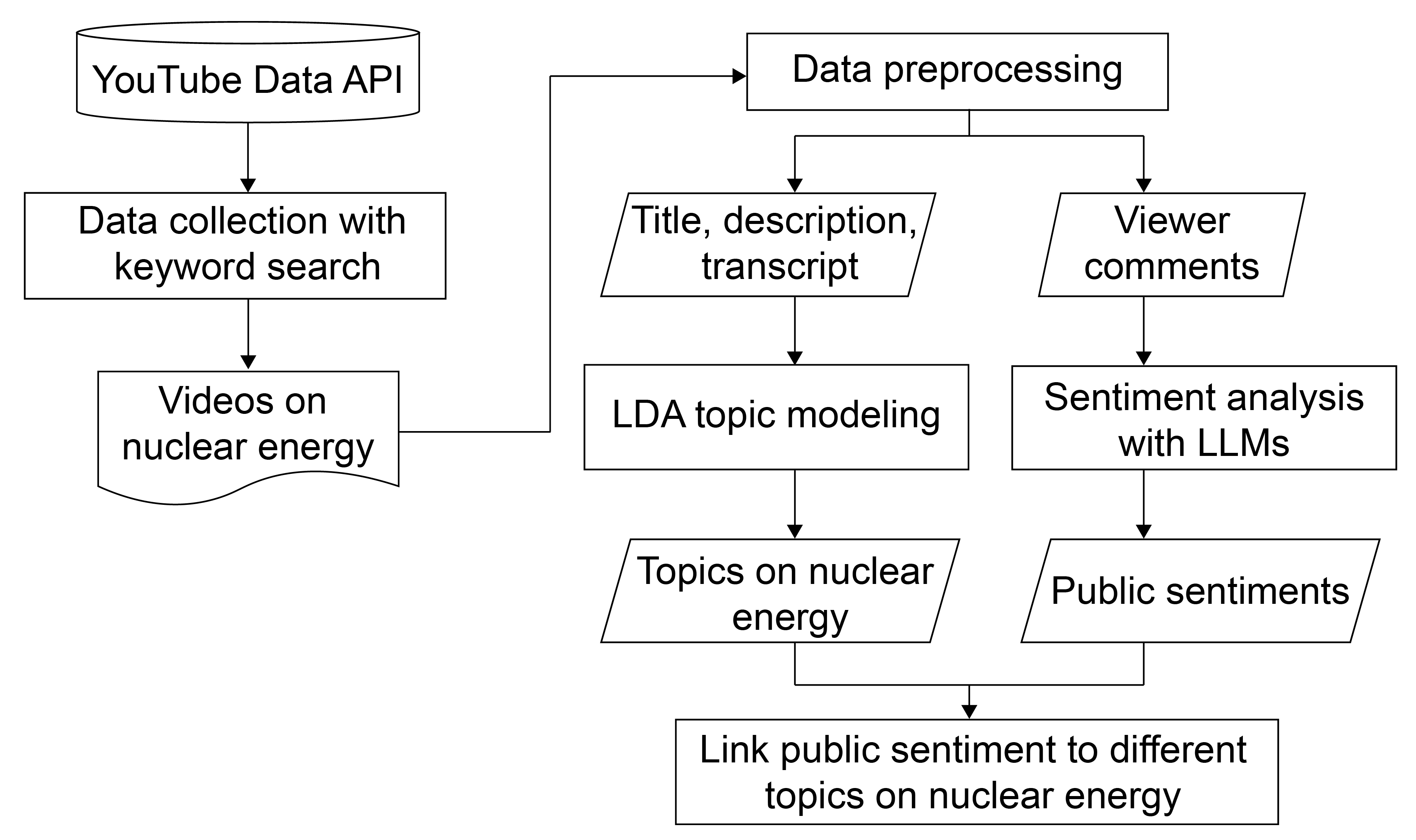}
    \caption{Flowchart illustrating the process to determine public sentiment towards various issues in nuclear energy media online. The process involves topic modeling of video content to identify key themes, followed by sentiment analysis of comments to evaluate public sentiment towards each identified topic.}
    \label{fig:workflow}
\end{figure}

To answer this question, we combined Latent Dirichlet Allocation (LDA) topic modeling with sentiment analysis to investigate nuclear energy-related content on YouTube, as illustrated in the flowchart in Figure \ref{fig:workflow}. First, LDA was applied to 3,101 YouTube videos published by official Japanese broadcasting stations to identify prominent themes. Then, we benchmarked the performance of BERT and GPT models on sentiment analysis using a curated test set, selecting the best-performing model to classify user sentiments. This combined approach directly links user sentiment to the main topics of the associated videos, achieving a level of topic-sentiment granularity not attained in previous studies. As a result, this work not only uncovers the online content landscape surrounding nuclear energy but also offers a deeper understanding of public perception of specific issues, which is crucial for the continued development of nuclear energy in Japan.

\section{Data collection and processing}
Data on YouTube videos, including titles, descriptions, comments, and metrics, were extracted using the YouTube Data API. The search was conducted using the keywords "\begin{CJK}{UTF8}{ipxm}福島第一原発\end{CJK}" (Fukushima Daiichi nuclear power plant) or "\begin{CJK}{UTF8}{ipxm}原子力\end{CJK}" (nuclear energy)  up to April 1, 2024. To ensure the quality and reliability of the video content, we retained 7,505 videos from the 15 major Japanese broadcasting stations. Additionally, the automatically generated Japanese subtitles of these videos were obtained using the YouTube Transcript API \cite{youtube-transcript-api}, providing additional textual context necessary for higher-quality topic modeling. Elements such as URLs, hashtags, and their associated texts were removed from the collected data.

\begin{figure}[h]
    \centering
    \includegraphics[width=0.7\linewidth]{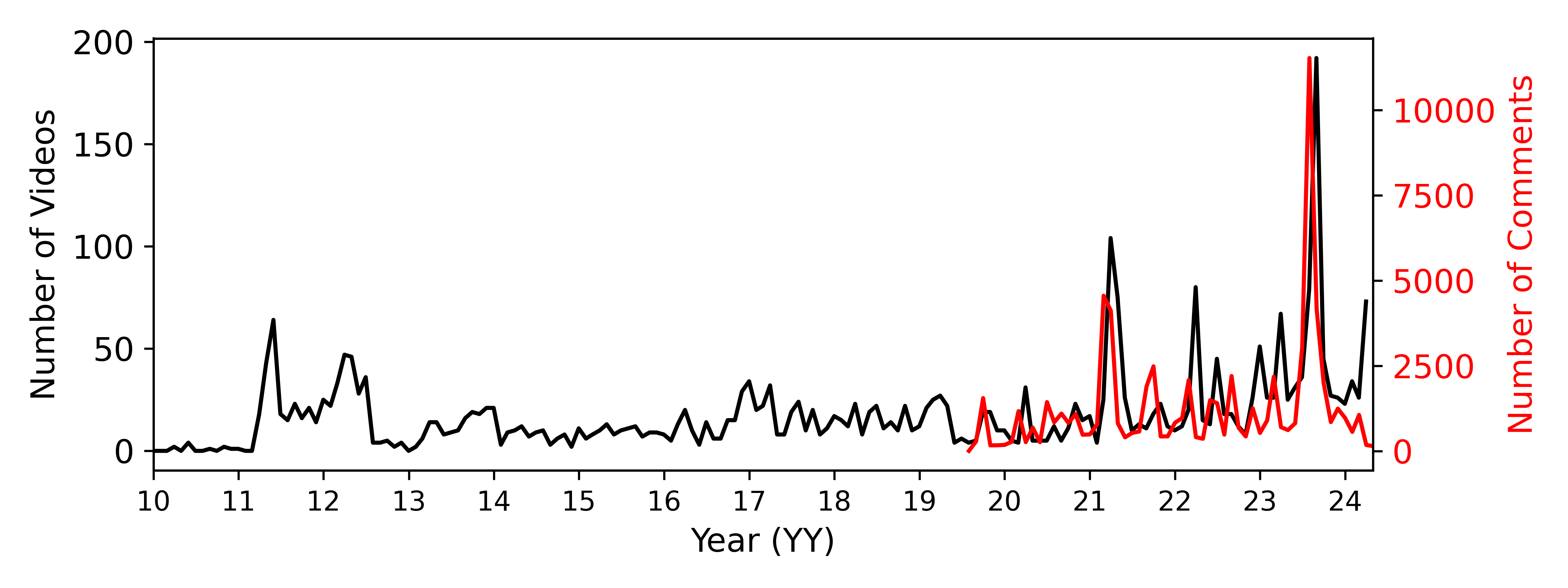}
    \caption{Monthly distribution of collected videos and viewer comments}
    \label{fig:time-series-video}
\end{figure}

To ensure the collected videos focused on domestic issues related to nuclear energy, we applied a rule-based filter to remove videos lacking "\begin{CJK}{UTF8}{ipxm}原発\end{CJK}" (nuclear power stations) or "\begin{CJK}{UTF8}{ipxm}原子力\end{CJK}" (nuclear energy) in their title or description, as well as those mentioning country names such as "North Korea" or "Russia," to minimize the influence of international politics on user sentiment, which is outside the scope of this study. This filtering process resulted in a final dataset of 3,101 videos in text format, with their time-series distribution shown in Figure \ref{fig:time-series-video}. For clarity, we refer to the textual content of these videos, including titles, descriptions, and subtitles, as the corpus.

Subsequently, as the goal of this work is to investigate Japanese citizens' sentiment on nuclear energy, we removed non-Japanese comments using a fine-tuned RoBERTa model for language identification tasks \cite{luca_papariello_2024}. The model's performance in identifying Japanese and non-Japanese comments was tested on a set of 500 human-annotated comments from this study, achieving an excellent accuracy of 0.992. Among the 74,699 text-mined comments from the 3,101 videos, the RoBERTa model labeled 2,021 as "not Japanese," which were subsequently removed. The time-series distribution of the remaining 72,678 comments is also shown in Figure \ref{fig:time-series-video}, where the lack of user comments before late 2019 is attributed either to low levels of user engagement or to the comment sections being turned off during that period.

\section{Methodology}
\label{sec:method}
\subsection{Topic modeling}
\label{subsec:topic model}
\begin{figure}[h]
    \centering
    \includegraphics[width=0.35\linewidth]{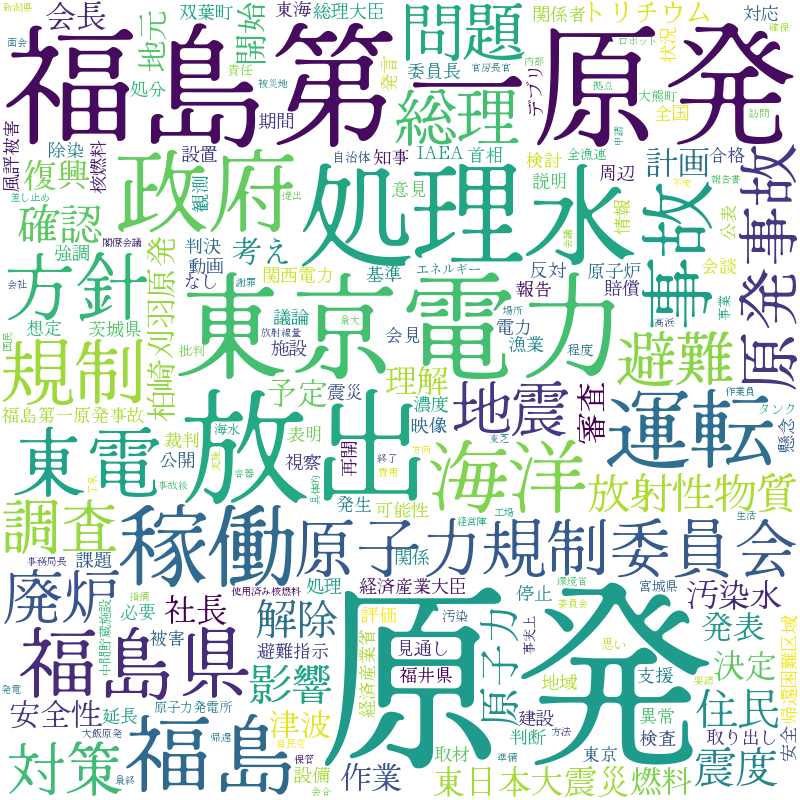}
    \caption{Word cloud generated from the bag-of-words vectors.}
    \label{fig:bow_dictionary}
\end{figure}

Prior to topic modeling, we tokenized the corpus and converted the contents of each video into a bag-of-words (BoW) representation, focusing on semantically important nouns that appeared at least once. This approach has been shown to improve topic modeling quality by reducing noise from irrelevant and uncommon words \cite{martin2015more,may2019analysislemmatizationtopicmodels}. Tokenization was performed using the Japanese morphological analyzer Sudachi \cite{sudachi}, chosen for its granular control over token generation and normalization features. Specifically, Sudachi's split mode C was used to produce the longest possible meaningful tokens, ensuring that words like \begin{CJK}{UTF8}{ipxm}福島第一原発\end{CJK} (Fukushima Daiichi Nuclear Power Station) were not split into \begin{CJK}{UTF8}{ipxm}福島, 第, 一, 原発\end{CJK}, preserving their original meaning. Finally, a word cloud generated from the BoW vectors is shown in Figure \ref{fig:bow_dictionary}.

Subsequently, we applied LDA \cite{blei2003latent} on the BoW vectors of each nuclear energy-related video for topic modeling. LDA learns underlying topics in the corpus by assuming that each document is a mixture of topics and that each topic consists of a set of frequently co-occurring words. Consequently, the meaning of each topic must be manually interpreted based on the top keywords assigned by the model. Additionally, as an unsupervised machine learning method, LDA does not automatically determine the optimal number of topics, which is typically inferred indirectly using metrics that assess topic quality, such as perplexity, divergence, or coherence scores. In this work, we prioritized coherence scores \cite{roder2015exploring} to determine the optimal number of topics, as higher coherence reflects stronger semantic similarity among the top words in each topic, which is generally associated with better topic interpretability.

\subsection{Sentiment analysis}
Sentiment analysis (SA), also commonly known as opinion mining, is the task of identifying the emotions, opinions, or tone within a given piece of information. In this work, given the large number of user comments collected, manual sentiment labeling is impractical. Therefore, we explored both lexicon-based and LLM-based SA methods, comparing their performance using a set of 500 human-annotated comments. The best-performing method was then used to process the 72,678 user comments.

\subsubsection{Lexicon-based approach}
While the performance of lexicon-based SA methods is inherently limited by the static nature of sentiment dictionaries, we implemented this approach due to its simplicity compared to LLMs. Lexicons require no training or fine-tuning with large amounts of human-annotated data, and each word is assigned a straightforward sentiment value. In this work, we utilized the oseti \cite{oseti} Python library to calculate the sentiment score of each comment based on the total number of positive and negative words as follows:

\begin{equation}
    \textrm{Sentiment score} = \frac{\textrm{\# positive - \# negative}}{\textrm{\# positive + \# negative}}
\end{equation}

The sentiment (positive or negative) of each word is determined using two human-annotated sentiment dictionaries \cite{dict1, dict2}. Comments with sentiment scores between -0.33 and 0.33 are labeled as neutral, while those above 0.33 are labeled as positive, and those below -0.33 as negative.

\subsubsection{LLM-based approach}
Unlike lexicons, LLMs based on the transformer architecture \cite{vaswani2023attentionneed} excel at considering contexts and understanding nuances in human speech. This is especially important for accurately identifying the sentiment in social media comments, where informal language and irony are prevalent. To select the best-performing SA model for our dataset, we evaluated and compared the performance of both BERT and GPT models on an annotated test set.

For BERT-based models, SA is a downstream task fine-tuned using the embedding of the model's [CLS] token from its output layer. Consequently, the SA performance of BERT models depends heavily on the data used during fine-tuning. In this work, we tested two Japanese BERT models fine-tuned for SA: one trained on an unspecified dataset (referred to as BERT-koheiduck) \cite{koheiduck-bert} and another trained on Amazon user reviews (referred to as BERT-phu) \cite{christian-phu-bert}.

In contrast, GPT leverages its language comprehension and generative capabilities to perform SA directly through prompting, without requiring fine-tuning. Here, we evaluated the latest GPT-4o-2024-08-06 model on SA tasks using both zero-shot and few-shot approaches. In zero-shot prompting, GPT performs sentiment analysis without prior examples, whereas in the few-shot approach, several examples are included in the prompt to guide its responses \cite{brown2020languagemodelsfewshotlearners}. In this work, we included six examples, two for each sentiment class, in our few-shot prompt. The prompts used in this study, as shown in Table \ref{tab:gpt_prompts}, were designed to clearly define the SA task and output format \cite{zhang2023sentimentanalysiseralarge}. In addition, we set a random seed \cite{krugmann2024sentiment} and fixed the "temperature" parameter to 0 \cite{kheiri2023sentimentgptexploitinggptadvanced,zhang2023sentimentanalysiseralarge,krugmann2024sentiment} to enhance the reproducibility of GPT's response.

\begin{table}[h]
	\caption{Zero-shot and few-shot prompts used for sentiment analysis. Only one of the six examples included in the few-shot prompt is shown here for brevity.}
	\centering
	\begin{tabular}{p{0.1\linewidth} p{0.8\linewidth}}
    \toprule
    \textbf{System (zero-shot)} & 
    \begin{CJK}{UTF8}{ipxm}
    次に提供されるYouTubeのコメントの全体的な感情を分類してください。感情の分類は、以下の3つの選択肢から1つを選んでください：ポジティブ、中立・判断不可能、ネガティブ。選択肢のみを出力してください。
    \end{CJK} \newline
    (Please classify the overall sentiment of the provided YouTube comment. Select one of the three following options as the sentiment classification: positive, neutral/indeterminate, negative. Only output the selected option.) \\
    \addlinespace
	\textbf{User} & (Comment for sentiment analysis) \\
    \midrule
    \textbf{System (few-shot)} & 
    \begin{CJK}{UTF8}{ipxm}
    次に提供されるYouTubeのコメントの全体的な感情を分類してください。感情の分類は、以下の3つの選択肢から1つを選んでください：ポジティブ、中立・判断不可能、ネガティブ。選択肢のみを出力してください。
    \end{CJK} \\     
    \addlinespace
    &
    \begin{CJK}{UTF8}{ipxm}
    例 (Example):
    \end{CJK} \\
    &
    \begin{CJK}{UTF8}{ipxm}
    コメント: しっかり勉強するのは良い事です自分を学び治す為にも (Comment: Studying hard is a good thing, as it helps you learn and improve yourself)  
    \end{CJK} \\
    &
    \begin{CJK}{UTF8}{ipxm}
    ポジティブ (Positive)
    \end{CJK} \\
    \addlinespace
	\textbf{User} & (Comment for sentiment analysis) \\
	\bottomrule
	\end{tabular}
	\label{tab:gpt_prompts}
\end{table}

\section{Results}
\subsection{Topic modeling}
\begin{figure}[h]
    \centering
    \includegraphics[width=0.4\linewidth]{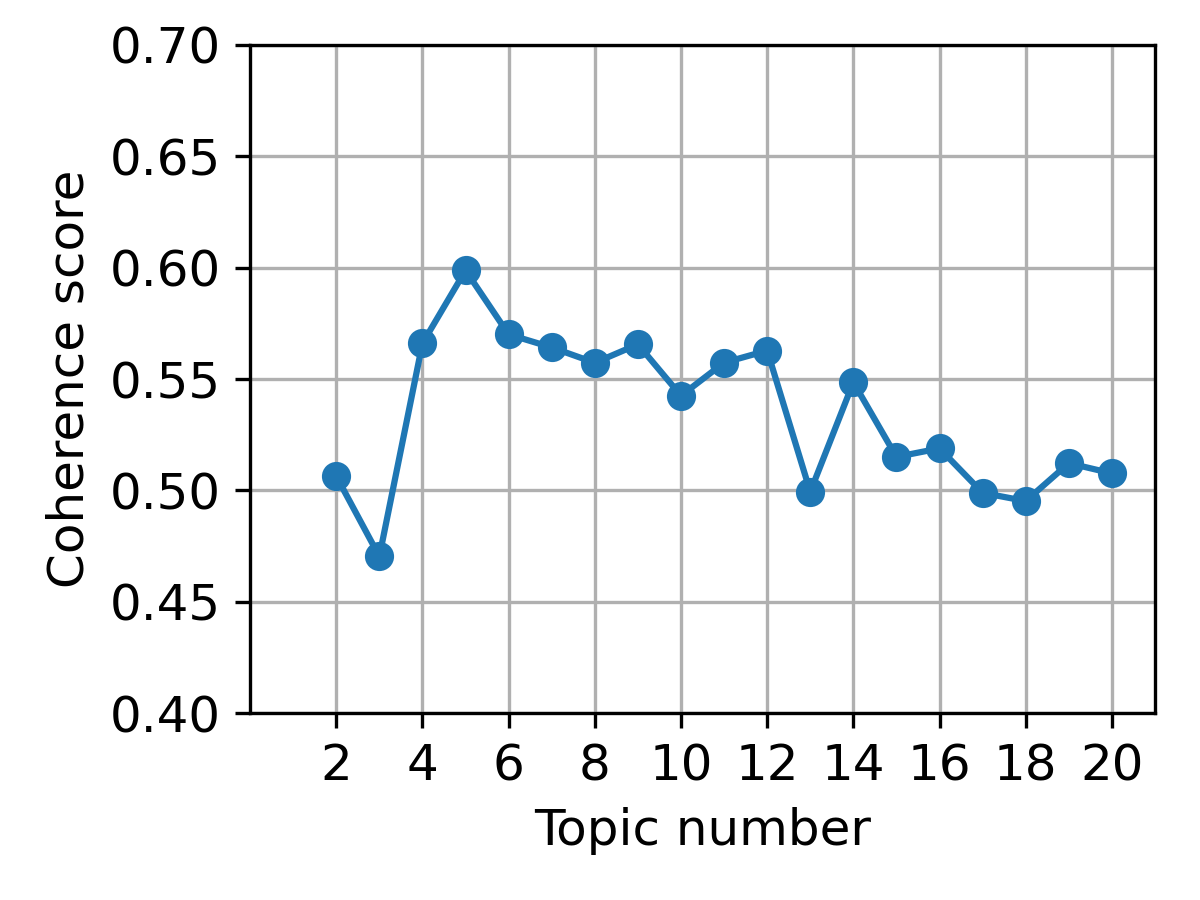}
    \caption{Relationship between topic number and coherence score of trained LDA models.}
    \label{fig:coherence-topic-number}
\end{figure}

As mentioned in Section \ref{subsec:topic model}, the LDA model's coherence score was used to evaluate the quality of generated topics and determine the optimal number of topics for the 3,101 videos. Accordingly, we trained 19 LDA models with topic numbers ranging from 2 to 20 and evaluated their coherence scores, as illustrated in Figure \ref{fig:coherence-topic-number}. 

The LDA model containing 5 topics achieved the highest score of 0.6, and coherence scores gradually decreased as the number of topics increased. However, closer inspection of the model's top keywords revealed that the small number of topics resulted in overly broad themes that overloaded individual topics with information. For instance, the model assigned words such as \begin{CJK}{UTF8}{ipxm}東京電力 (TEPCO), 原子力規制委員会 (Nuclear Regulation Authority), 柏崎刈羽原発 (Kashiwazaki-Kariwa Nuclear Power Plant), 福島第一原発 (Fukushima Daiichi Nuclear Power Plant), デブリ (debris), and 故障 (malfunction)\end{CJK} into a single topic. While the contents of these 5 topics can be easily interpreted, their broad scope lacked the specificity needed to analyze the online content landscape on nuclear energy effectively.

Consequently, to balance topic interpretability and granularity, we manually inspected the top keywords assigned to each topic across all trained LDA models, selecting the model that provided the most meaningful, distinct, and non-overlapping topics. This process led to the selection of the LDA model that categorized the contents of the 3,101 YouTube videos into 16 distinct topics. Detailed information on each topic, including the top 5 keywords and a human-interpreted topic phrase, is presented in Table \ref{tab:topic-phrase}. These identified topics reveal a diverse range of nuclear energy-related issues covered in news videos, including the Fukushima 1F accident (e.g., No.1 compensation, No.3 government response, No.5 treated water release, No.12 reactor decommissioning, and No.16 resident returns), the broader fuel cycle (e.g., No.7 interim storage and No.8 final disposal sites), No.6 energy policies, and other domestic nuclear plants, such as the Kashiwazaki-Kariwa (No.9) and Takahama (No.11) plants.

\begin{table}[h]
    \centering
    \caption{Human-interpreted themes and top key words of the 16 identified topics on domestic nuclear energy.}
    \begin{tabular}{p{0.03\linewidth}p{0.32\linewidth}p{0.55\linewidth}}
        \toprule
        \textbf{No.} & \textbf{Topic phrase} & \textbf{Top words} \\
        \toprule
        1 & Nuclear accident trials and compensation & \begin{CJK}{UTF8}{ipxm}事故 (accident), 判決 (verdict), 裁判 (trial), 避難 (evacuation), 原発事故 (nuclear accident)\end{CJK} \\
        \addlinespace
        2 & Extension of nuclear power plant operation period & \begin{CJK}{UTF8}{ipxm}運転 (operation), 期間 (duration), 原発 (nuclear power plant), 延長 (extension), 石川県 (Ishikawa Prefecture)\end{CJK} \\
        \addlinespace
        3 & Prime Minister and government responses, explanations, inspections & \begin{CJK}{UTF8}{ipxm}総理 (Prime Minister), 対応 (response), 政府 (government), 質問 (question), 視察 (inspection)\end{CJK} \\
        \addlinespace
        4 & Import and export of marine products from Fukushima and Tokai & \begin{CJK}{UTF8}{ipxm}東海 (Tokai), 輸入 (import), 海底 (seabed), 水産物 (marine products), 食品 (food)\end{CJK} \\
        \addlinespace
        5 & Release of treated water, impact on fisheries, reputational damage & \begin{CJK}{UTF8}{ipxm}放出 (release), 処理水 (treated water), 海洋 (ocean), トリチウム (tritium), 風評被害 (reputational damage)\end{CJK} \\
        \addlinespace
        6 & Energy policy, renewable energy, electricity, and nuclear power generation & \begin{CJK}{UTF8}{ipxm}原発 (nuclear power plant), エネルギー (energy), 電力 (electricity), 原子力 (nuclear power), 再生可能エネルギー (renewable energy)\end{CJK} \\
        \addlinespace
        7 & Spent nuclear fuel, interim storage facility, reprocessing plant & \begin{CJK}{UTF8}{ipxm}中間貯蔵施設 (interim storage facility), 使用済み核燃料 (spent nuclear fuel), 工場 (factory), 再処理 (reprocessing), 建設 (construction)\end{CJK} \\
        \addlinespace
        8 & Investigation and selection of final disposal site & \begin{CJK}{UTF8}{ipxm}調査 (investigation), 最終 (final), 処分 (disposal), 候補 (candidate), 最終処分場 (final disposal site)\end{CJK} \\
        \addlinespace
        9 & Nuclear Regulation Authority activities and issues with Kashiwazaki-Kariwa & \begin{CJK}{UTF8}{ipxm}規制 (regulation), 原子力規制委員会 (Nuclear Regulation Authority), 柏崎刈羽原発 (Kashiwazaki-Kariwa), 問題 (issue), 委員長 (Chair)\end{CJK} \\
        \addlinespace
        10 & Earthquakes and tsunamis, damage, and impact & \begin{CJK}{UTF8}{ipxm}地震 (earthquake), 震度 (seismic intensity), 観測 (observation), 情報 (information), 発生 (occurrence)\end{CJK} \\
        \addlinespace
        11 & Kansai Electric Power and Takahama Plant & \begin{CJK}{UTF8}{ipxm}関西電力 (Kansai Electric Power Co.), 発電 (power generation), 高浜 (Takahama), 発電所 (power station), 稼働 (operation)\end{CJK} \\
        \addlinespace
        12 & Decommissioning and contaminated water & \begin{CJK}{UTF8}{ipxm}作業 (work), 汚染水 (contaminated water), 分析 (analysis), 取り出し (extraction), 福島第一原発 (Fukushima Daiichi)\end{CJK} \\
        \addlinespace
        13 & Reflection and memories of the Great East Japan Earthquake disaster & \begin{CJK}{UTF8}{ipxm}震災 (disaster), 思い (thoughts), 自分 (self), 本当 (reality), 東日本大震災 (Great East Japan Earthquake)\end{CJK} \\
        \addlinespace
        14 & Footage of reactor containment and fuel debris & \begin{CJK}{UTF8}{ipxm}容器 (container), 格納 (containment), ロボット (robot), 映像 (footage), 公開 (release)\end{CJK} \\
        \addlinespace
        15 & Nuclear plant inspections, operations, and local approval & \begin{CJK}{UTF8}{ipxm}稼働 (operation), 知事 (governor), 原発 (nuclear plant), 同意 (agreement), 合格 (approval)\end{CJK} \\
        \addlinespace
        16 & Lifting evacuation orders, decontamination, and return of residents & \begin{CJK}{UTF8}{ipxm}解除 (lifting), 避難指示 (evacuation order), 双葉町 (Futaba Town), 除染 (decontamination), 福島県 (Fukushima Prefecture)\end{CJK} \\
        \bottomrule
    \end{tabular}
    \label{tab:topic-phrase}
\end{table}

\begin{figure}[h]
    \centering
    \includegraphics[width=\textwidth]{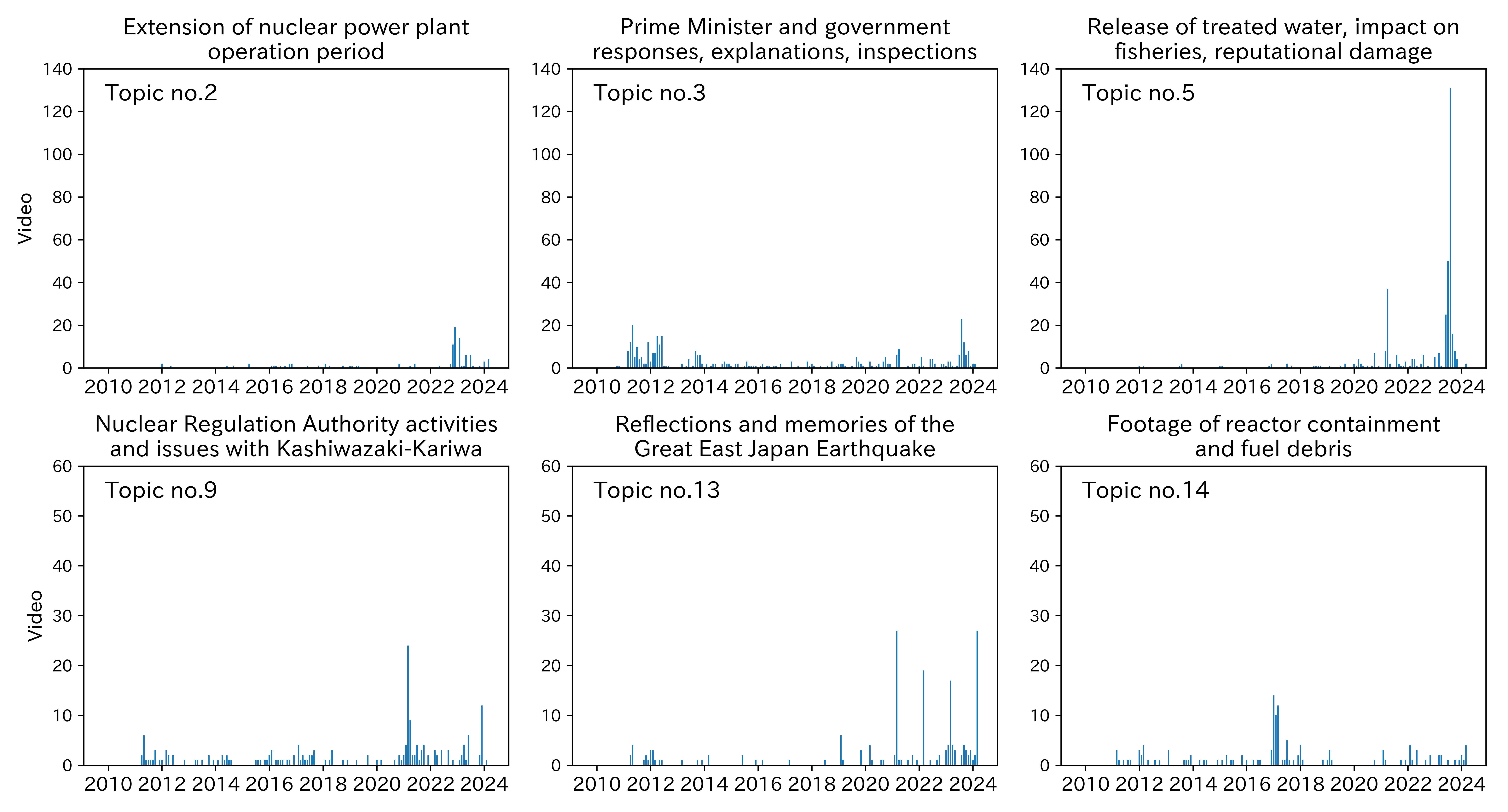}
    \caption{Number of monthly published videos of selected topics.}
    \label{fig:time-series-topic}
\end{figure}

To verify the validity of the human-interpreted topic phrases, we analyzed the time-series trends of several topics, as shown in Figure \ref{fig:time-series-topic}, to investigate their correspondence with real-world events. While LDA assigns multiple topics to each video, we simplified the topic assignment by designating only a single "main topic" for each video to facilitate sentiment analysis. Specifically, having multiple topics in a single video makes it challenging to determine which topic a user’s sentiment is addressing. Consequently, the "main topic" for each video was selected as the one with the highest word count in the document-topic distribution.

From Figure \ref{fig:time-series-topic}, we can observe excellent alignments between spikes in the number of videos for specific topics and the timing of relevant real-world events. For example, the two major peaks for videos related to Topic No.5 "release of treated water" occurred in 2021-04 and 2023-08, corresponding to the government’s announcement of plans to release treated water and the initial release, respectively. Similarly, peaks in videos about Topic No.9 "nuclear regulation authority" coincided with the suspension of Kashiwazaki-Kariwa nuclear plant's operations in 2021-03 and their resumption in 2023-12.  A detailed list of real-world events corresponding to these peaks is presented in Table \ref{tab:spikes-explained}, confirming the validity of our topic modeling approach and the human-interpreted topic phrases.

\begin{table}[h]
    \caption{Major spikes in topic coverage by online news videos and related real-world events.}
    \centering
    \begin{tabular}{>{\raggedright\arraybackslash}p{0.4\linewidth}cc}
    \toprule
    \textbf{Topic} & \textbf{Spike} & \textbf{Related event} \\ 
    \midrule
      \multirow{2}{=}{\textbf{No.2} Extension of nuclear power plant operation period} 
         & 2022/12 & Regulatory framework on extension approved  \\   
         & 2023/02 & Official decision for reactor operation extension  \\        
    \addlinespace
      \multirow{2}{=}{\textbf{No.3} Prime Minister and government responses, explanations, inspections} 
         & 2011/05 & Response to the Fukushima 1F accident \\   
         & 2023/08 & Comments on treated water release \\
    \addlinespace
      \multirow{2}{=}{\textbf{No.5} Release of treated water, impact on fisheries, reputational damage} 
         & 2021/04 & Decision to release treated water \\
         & 2023/08 & Initial release of treated water \\
    \addlinespace
      \multirow{2}{=}{\textbf{No.9} Nuclear Regulation Authority activities, Kashiwazaki-Kariwa issues} 
         & 2021/03 & Suspension of Kashiwazaki-Kariwa \\   
         & 2023/12 & Resuming operations at Kashiwazaki-Kariwa \\
    \addlinespace
      \multirow{2}{=}{\textbf{No.13} Reflections and memories of the Great East Japan Earthquake} 
         & \multirow{2}{*}{Every March} &  \multirow{2}{*}{Anniversaries of the Great East Japan Earthquake} \\ 
         & & \\  
    \addlinespace
      \multirow{2}{=}{\textbf{No.14} Footage of reactor containment and fuel debris} 
         & 2017/01 & Footage of Fukushima Daiichi Unit 2 \\   
         & 2017/03 & Footage of Fukushima Daiichi Unit 1 \\      
    \bottomrule   
    \end{tabular}
    \label{tab:spikes-explained}
\end{table}

\subsection{Sentiment analysis}
\subsubsection{Model benchmark results}
\begin{figure}[h]
    \centering
    \includegraphics[width=\linewidth]{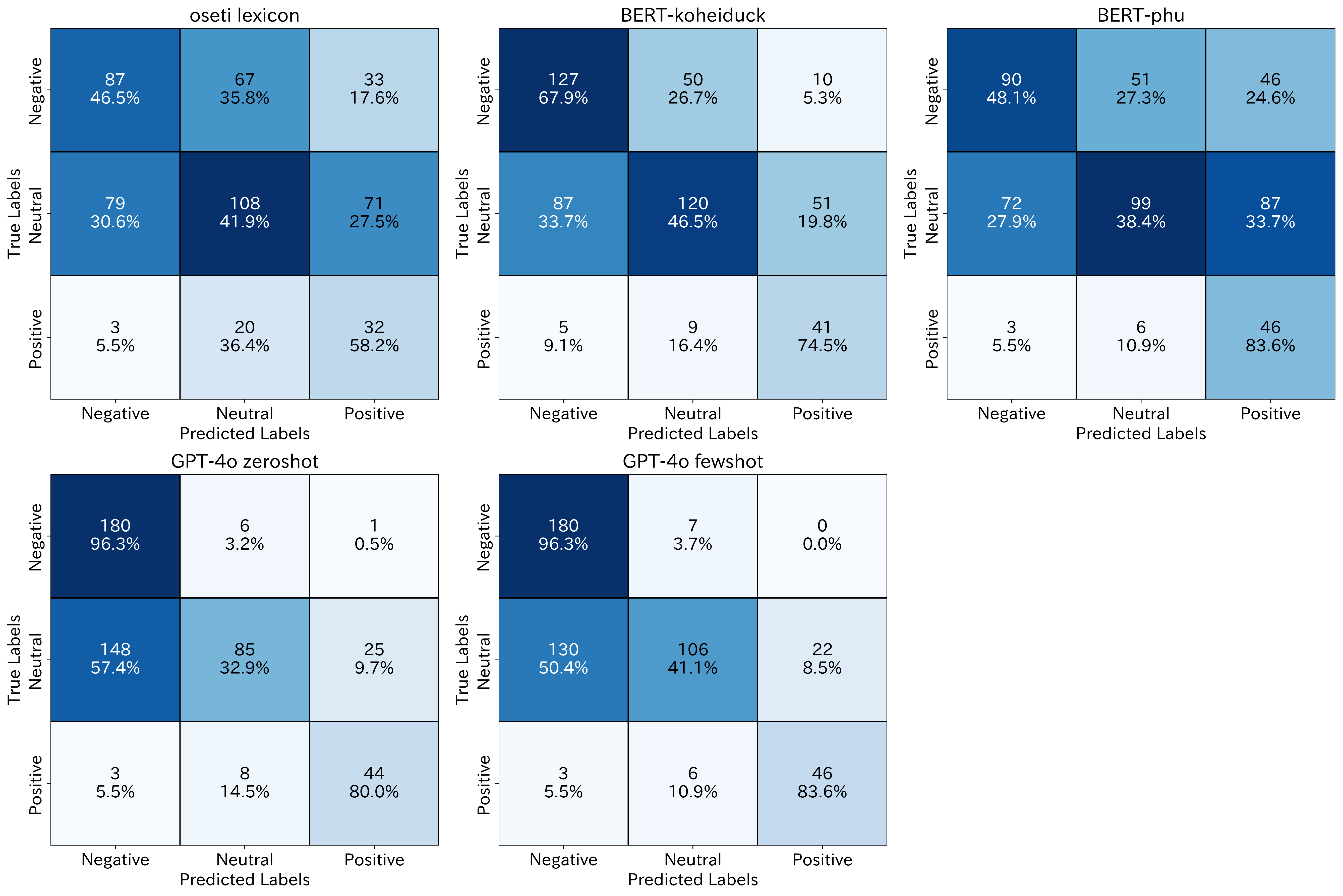}
    \caption{Confusion matrices of different sentiment analysis models on the benchmark set.}
    \label{fig:model-SA-benchmark-result}
\end{figure}

As outlined in Section \ref{sec:method}, lexicon-based and LLM-based SA methods were employed to classify the sentiment of user comments, given the large volume of collected data. To evaluate the performance of these models (lexicon, BERT, and GPT-4o) on SA tasks, we benchmarked them against a dataset of 500 YouTube comments, representative of the entire collected comments dataset. Each comment's sentiment was independently annotated by five native Japanese speakers from Kyoto University as positive, neutral/indeterminate, or negative. Ground-truth labels were assigned based on majority vote to ensure the reliability of the benchmark dataset. The benchmarked performance of each SA method is displayed in Figure \ref{fig:model-SA-benchmark-result} and Table \ref{tab:benchmark-results}. The metrics in Table \ref{tab:benchmark-results} are all weighted averages that account for data imbalance among the three sentiment classes, providing a more accurate measurement of each model's overall performance on the benchmark set and the collected comments.

\begin{table}[h]
    \centering
    \caption{Weighted performance metrics of different sentiment analysis models on the benchmark set.}
    \begin{tabular}{lcccc}
    \toprule
    \textbf{Method} & \textbf{Accuracy} & \textbf{Precision} & \textbf{Recall} & \textbf{F1 Score} \\
    \midrule
    oseti lexicon & 0.45 & 0.50 & 0.45 & 0.47 \\
    BERT-koheiduck & 0.58 & 0.61 & 0.58 & 0.57 \\
    BERT-phu & 0.47 & 0.56 & 0.47 & 0.48 \\
    GPT-4o zeroshot & 0.62 & 0.72 & 0.62 & 0.58 \\
    \textbf{GPT-4o fewshot} & \textbf{0.66} & \textbf{0.75} & \textbf{0.66} & \textbf{0.64} \\
    \bottomrule
    \end{tabular}
    \label{tab:benchmark-results}
\end{table}

From the benchmark results, it is evident that the lexicon-based approach using the oseti \cite{oseti} Python library performed the worst, with an accuracy of only 0.45. Its poor performance is attributed to the static sentiment values assigned to each word, which struggle to identify irony and sarcasm prevalent in short-format online comments. Compared to LLM-based approaches, oseti tends to label sentiments as neutral, likely due to both the lack of recorded sentiment values for newer words and outdated sentiment values assigned to words whose connotations have evolved over time.

Among the two fine-tuned BERT models, BERT-phu \cite{christian-phu-bert}, which was fine-tuned on Amazon reviews, displayed only marginally better performance than oseti. Its struggle to classify the sentiment of short-format YouTube comments is likely due to the stark contrast between the context of Amazon reviews and YouTube comments. In contrast, the BERT-koheiduck model \cite{koheiduck-bert} achieved better performance, with an accuracy of 0.58. However, since the dataset and data source used to fine-tune BERT-koheiduck are unknown, the reasons for the performance differences between the two models remain unclear. One shared limitation of the two BERT models was the misclassification of approximately 30$\%$ of negative comments as neutral. This issue was significantly mitigated by the GPT-4o model, which correctly identified over 96$\%$ of the negative comments in the benchmark set, outperforming BERT-based models irrespective of prompting strategy. Between the two prompting strategies for GPT, few-shot prompting outperformed zero-shot prompting across all three sentiment classes (positive, neutral, and negative), consistent with findings from previous studies \cite{zhong2023chatgptunderstandtoocomparative,krugmann2024sentiment}. 

Interestingly, while GPT-4o correctly classified Japanese comments with negative and positive sentiments with over 80$\%$ accuracy, it exhibited a surprising pessimism toward comments considered neutral or indeterminate by human annotators, frequently labeling them as negative. While few-shot prompting reduced this tendency, GPT-4o still misclassified 50.4$\%$ of neutral comments as negative. Similar biases have been observed in older models, such as GPT-3.5 and Llama 2, where models tended to be overly optimistic. These biases are speculated to result from influences during the model's training and the characteristics of the dataset used for sentiment analysis \cite{krugmann2024sentiment}. 

In the context of this work, we speculate that part of this pessimism stems from differing thresholds for distinguishing between negative and neutral sentiments in humans and GPT-4o, based on the misclassified comments. GPT-4o likely requires less context before making a decision, leading it to mislabel short-format comments such as "So who will take responsibility?" and "If there's no news coverage, there won't be any rumors" as negative, whereas human annotators considered them neutral or indeterminate. While such behavior could potentially be adjusted with further prompt engineering, a full examination of GPT-4o’s performance is beyond the scope of this work. Finally, as shown in Table \ref{tab:benchmark-results}, among the five SA methods evaluated, GPT-4o with few-shot achieved the highest overall accuracy of 0.66. Consequently, despite its slightly pessimistic tendencies, it was selected to evaluate the sentiment of over 70,000 user comments on videos related to nuclear energy.

\subsubsection{Sentiment analysis results}
\begin{figure}[h]
    \centering
    \includegraphics[width=0.7\linewidth]{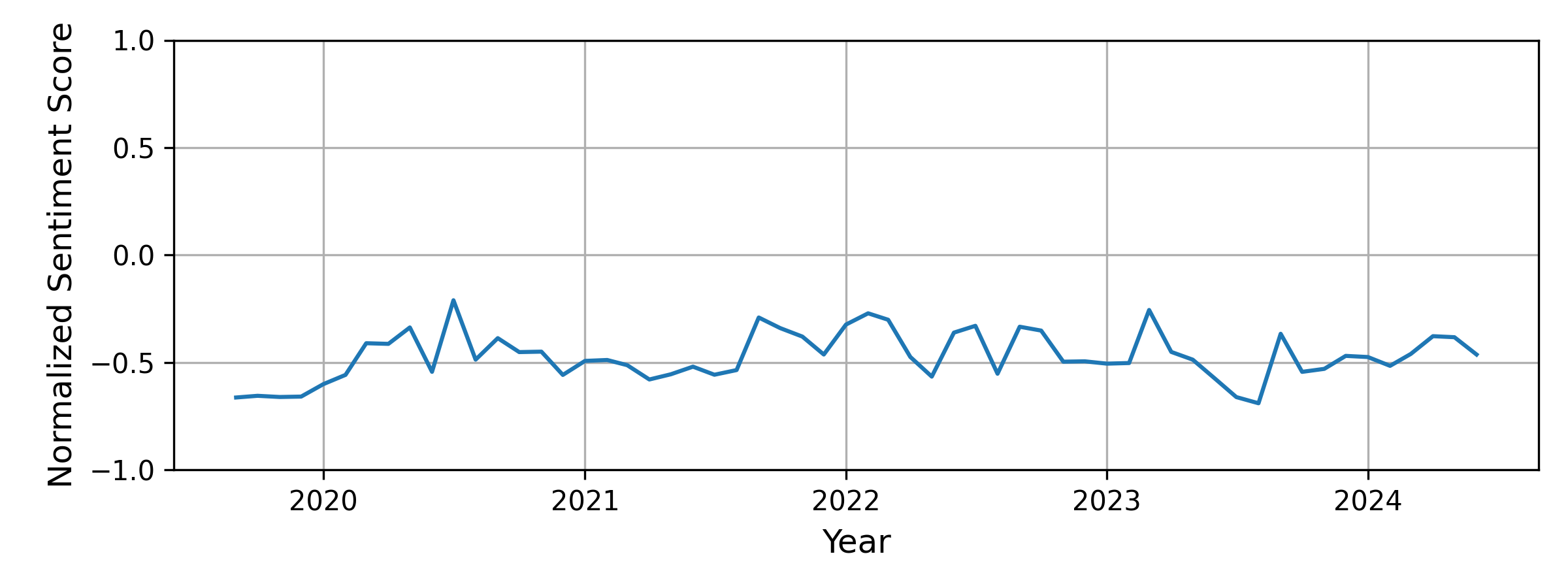}
    \caption{Normalized monthly sentiment scores across all topics.}
    \label{fig:normalized sentiment score}
\end{figure}

Using the sentiment assigned to each comment by the GPT-4o model, we calculated a monthly sentiment score across all topics, as shown in Figure \ref{fig:normalized sentiment score}. Although official news videos on nuclear energy have been published online since 2010, sentiment analysis is conducted on comments made after late 2019, when user engagement became more substantial. Here, positive comments were assigned a score of 1, neutral or intermediate comments 0, and negative comments -1, with the total score normalized by the number of comments per month. For example, a monthly normalized sentiment score of 1 indicates that all comments for a given month were positive. 

As shown in Figure \ref{fig:normalized sentiment score}, the overall online sentiment towards domestic nuclear energy-related issues remained relatively negative, with a sentiment score of approximately -0.5, without showing further decline over time. However, it is important to keep in mind the GPT-4o model's pessimistic tendencies, frequently assigning neutral comments a negative sentiment, as revealed by the benchmark results. Therefore, it is highly likely that while the overall tone remains in the negative range, it is closer to neutral, with the negativity exaggerated by the model.

While a direct comparison between our results and previous sentiment analysis studies on nuclear energy is challenging due to differences in the data source, region, and time frame, all studies point to a negative sentiment in the online discourse surrounding nuclear energy. For example, a domestic study by Hasegawa et al. \cite{hasegawa2020changing} on Japanese tweets related to radiation from 2011 to 2012 demonstrated a negative sentiment score, largely attributed to the aftermath of the Fukushima 1F accident, despite trending positively over time. Similarly, international studies have also indicated a generally negative online sentiment associated with nuclear energy or policy, often attributed not only to the Fukushima 1F accident but also to various domestic issues \cite{park2019positive,jeong2021sentiment} and international political affairs \cite{zarrabeitia2023nuclear,kwon2024sentiment}.

\begin{figure}[h]
    \centering
    \includegraphics[width=\linewidth]{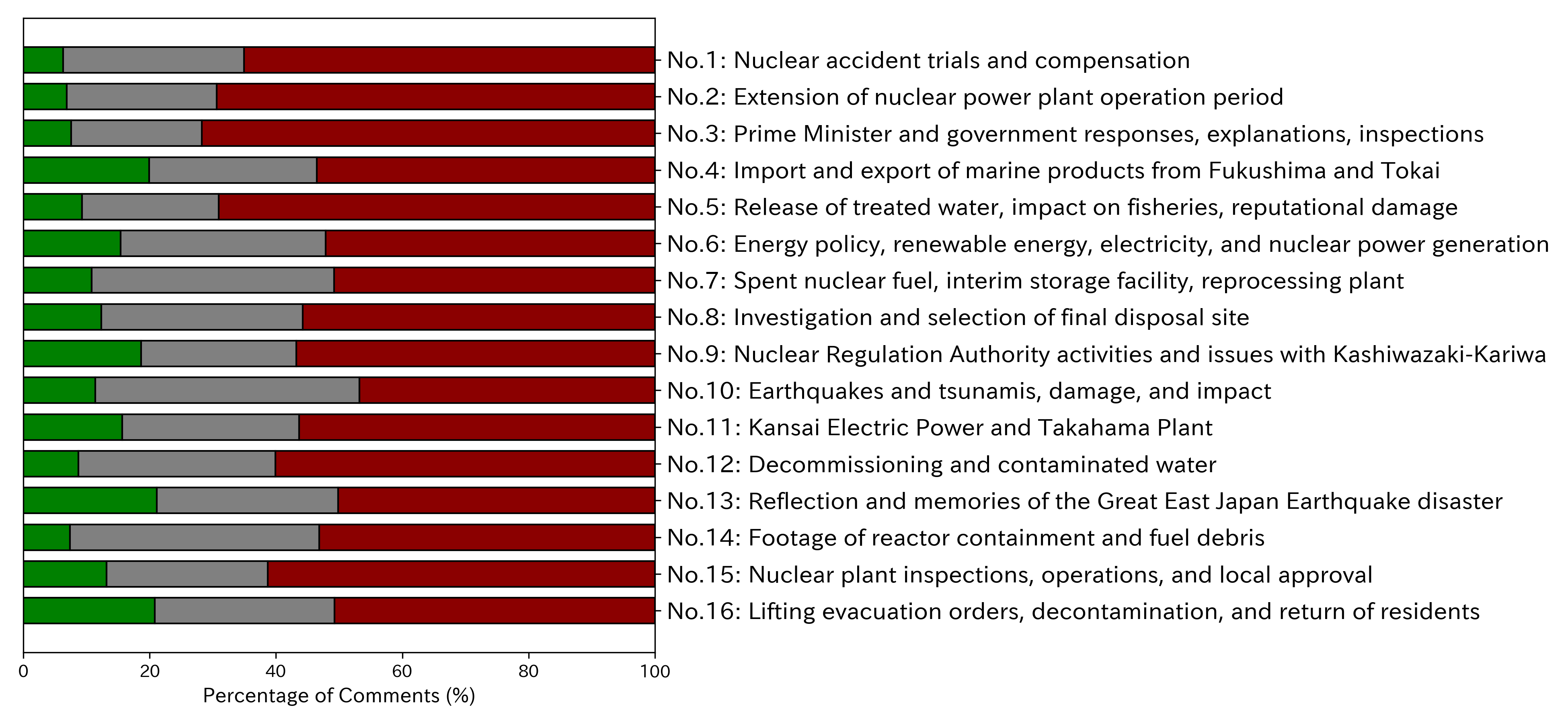}
    \caption{Percentage of comments assigned each sentiment for all 16 topics. Green: positive, Gray: neutral or indeterminate, and Red: negative.}
    \label{fig:percent of comments each sentiment}
\end{figure}

To gain more insight into the sentiment towards each respective topic and control for the model's negative bias, we compared the sentiment distribution across the 16 extracted topics in Figure \ref{fig:percent of comments each sentiment}, where green, gray, and red represents the share of positive, neutral, and negative comments, respectively. It is evident that news videos focusing on "reflection and memories of the Great East Japan Earthquake" (Topic No.13), "lifting of evacuation orders" (Topic No.16), and "import and export of marine products" (Topic No.4) invoked the most positive response from the viewers, as these topics generally reflect improvement of conditions after the Fukushima 1F accident.

In contrast, the topics with the highest percentages of negative comments are "responses, explanations from the government" (Topic No.3) and "release of treated water" (Topic No.5), followed by "extension of nuclear plant operation periods" (Topic No.2). While clarifying the motivations behind these sentiments would require further investigation into the comment contents, they are likely influenced by the public's difficulty in fully grasping and internalizing the large volumes of information on these issues. For example, a 2022 poll by the Japan Atomic Energy Relations Organization revealed that approximately 30$\%$ of respondents had never heard of the terms related to treated water included in the poll, and more than 80$\%$ lacked a clear understanding to explain them \cite{jaero2022}. Therefore, while online comments on "release of treated water" (Topic No.5) display an overwhelming amount of negative sentiment, this sentiment is likely influenced by a knowledge gap between the public and the government, as well as personal emotions.

\subsection{Co-occurrence networks analysis}
To investigate the details of viewer comments, we conducted word co-occurrence network analysis by identifying word pairs that frequently appear together in the same sentence. Among the 16 topics identified in this work, we specifically focused on clarifying how the content of viewer comments on news videos discussing "release of treated water, impact on fisheries, reputation damage" changed before and after the initial treated water release date on August 24th, 2023. This focus was partly motivated by the large volume of comments available during this period, as shown in Figure \ref{fig:time-series-video}, allowing for a more comprehensive investigation.

\begin{figure}[h]
    \centering
    \includegraphics[width=\linewidth]{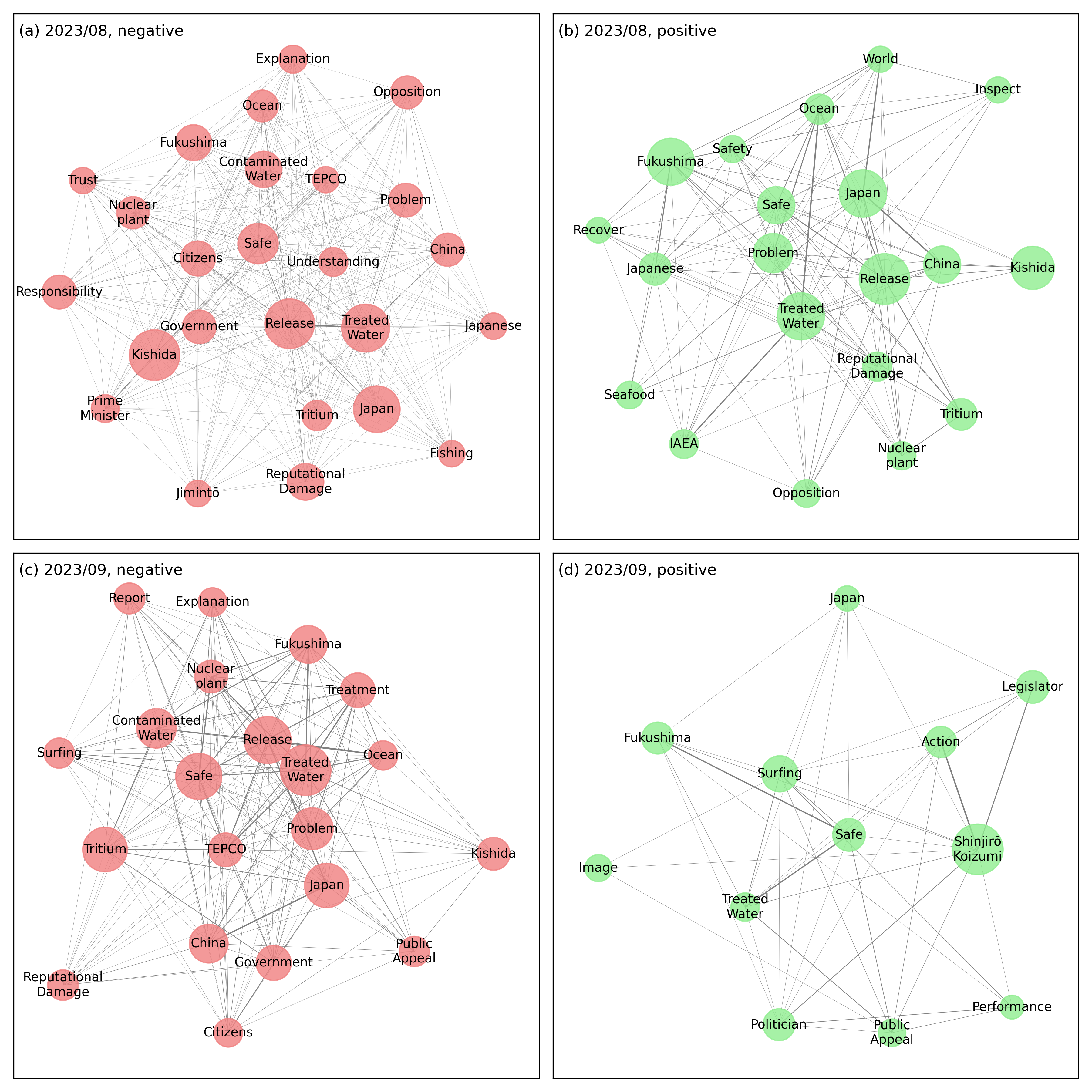}
    \caption{Word co-occurrence networks constructed from negative and positive comments associated with videos on "release of treated water," "impact on fisheries," and "reputational damage" during August and September 2023. The original Japanese terms have been translated into English.}
    \label{fig:co-occurence-networks}
\end{figure}

Consequently, we constructed four word co-occurrence networks based on the positive and negative comments in August and September of 2023, respectively, as shown in Figure \ref{fig:co-occurence-networks}. Here, each word's node size correlates to its appearance frequency, and the edge width represents the frequency a specific word pair appears together in the same sentence. To ensure readability of the co-occurrence networks, we reduced the number of nodes by showing only nouns that appeared above a certain threshold and translated the original Japanese terms into English.

As illustrated in Figure \ref{fig:co-occurence-networks}, the four co-occurrence networks share common nodes like "Reputational Damage," "Fukushima," "Ocean," and "Treated Water." These shared nodes indicate that viewer comments are focused on the issue of "treated water release" rather than unrelated topics. However, as these common nodes do not reveal the specific arguments behind each sentiment, it is essential to identify unique nodes that uncover the reasoning behind positive and negative comments.

A comparison between the networks for positive and negative comments reveals that political terms, such as "Jiminto," "Government," "Prime Minister", and institutional references like "TEPCO" appear predominantly in negative comments. This suggests that online comments with negative sentiments about the release of treated water may be politically motivated rather than focused solely on the treated water itself. Additionally, from August to September 2023, the structure and nodes of the co-occurrence networks for negative comments showed little change, indicating that the content of negative online discussions remained largely consistent.

In contrast, unique nodes such as "Recover" and "IAEA" emerged in the network for positive comments during August 2023, as shown in Figure \ref{fig:co-occurence-networks}(c). These nodes indicate that, along with expressions of support for the recovery from the Fukushima 1F accident, many positive comments referenced the IAEA. The frequent mentions of the IAEA in these positive comments suggest that its statements may have improved public perception of the treated water release, contributing to more favorable sentiments. This simultaneously highlights the important role of unbiased international third parties in shaping pubic understanding of complicated issues. 

Interestingly, the content of positive comments shifted dramatically in September 2023, the month following the initial release of the treated water. As shown in Figure \ref{fig:co-occurence-networks}(d), the focus of positive comments turned toward praising the actions of the Japanese politician Shinjiro Koizumi, who went surfing off the coast of Fukushima to demonstrate the safety of the treated water to the public. While it is challenging to determine whether these positive comments are directed more toward Koizumi himself or the treated water, this shift indicates that the actions by public figures, such as politicians, can significantly influence online public discourse. A similar case, where public figure created new topics for online discussion, was observed in a previous study by Yagahara et al. \cite{yagahara2018relationships} on Japanese Tweets posted in the immediate aftermath of the Fukushima 1F accident.

\section{Limitations and Future Work}
While this study demonstrated the effectiveness of natural language processing in extracting valuable insights from large volumes of online data related to nuclear energy, certain aspects of the current approach can be further refined in future studies.

\paragraph{User Demographic} Unlike public surveys that strive to gather a representative sample of the population, the characteristics of the online demographic cannot be controlled and likely differ from those of the general public. For example, opinions on Twitter have been shown to not necessarily correspond to those obtained from national surveys, due to the dynamic nature of its users \cite{mitchell2013twitter}. This potential mismatch has been a consistent challenge for past studies on online opinions about nuclear energy \cite{li2016tweeting,park2019positive,jeong2021sentiment,zarrabeitia2023nuclear}, also limiting the applicability of our findings to the portion of the public vocal online about nuclear energy-related issues.

\paragraph{Data Quality} To fully represent the contents of each video for topic modeling, we leveraged each video's transcripts in addition to its title and description. However, as these transcripts are automatically generated by YouTube, ensuring their quality remains a challenge. While using LLMs like GPT to review these transcripts showed some potential, LLMs could not correct errors made during speech-to-text conversion. In future studies, ensuring high transcript quality through either manual review or more advanced speech-to-text algorithms will enable more accurate analysis of online video content.

\paragraph{Topics Granularity} Although we identified 16 distinct topics from over 3,000 collected YouTube videos on domestic nuclear energy issues, the identified topics remain broad and lack specificity. For example, the topic "Prime Minister and government response, inspections" does not indicate the specific issues or events to which the government responded. Consequently, to gain a deeper understanding of the detailed content within each topic, future work could complement topic modeling with knowledge graphs to reveal additional details within each topic.

\paragraph{Sentiment Categories} In this work, we classified online sentiment into positive, neutral or indeterminate, and negative. However, for complex issues related to nuclear energy, a more granular analysis could be more appropriate, as negative sentiment could encompass varied emotions such as disappointment, concern, or anger. Additionally, to explore potential drivers of certain sentiments, we used word co-occurrence analysis to identify specific word pairs associated with each sentiment. Future studies could incorporate aspect-based sentiment analysis to provide more detailed insights into the specific causes or targets of online sentiments.

\section{Conclusion}
In this work, we applied natural language processing techniques to analyze online coverage of domestic nuclear energy issues, aiming to understand the contents presented to the public and the corresponding online sentiments. Specifically, we processed over 3,000 YouTube videos from official news channels for topic modeling and conducted sentiment analysis on over 70,000 associated comments.

Using the LDA topic model, we identified 16 distinct topics on nuclear energy, ranging from "nuclear accident trials and compensation" to "lifting evacuation orders and the return of residents," all corresponding well to real-world events. For sentiment analysis, GPT-4o with few-shot prompting was employed to assess each comment's sentiment, revealing a consistently slightly negative sentiment towards nuclear energy in online discussions. Specifically, our results highlighted relatively higher negativity toward news coverage on "government response" and "release of treated water" among the 16 extracted topics. Further word co-occurrence analysis on comments related to "release of treated water" in August and September of 2023 suggests that a majority of the negative sentiments may be politically motivated. Additionally, statements from non-political organizations like the IAEA and actions by politicians were found to positively influence public perception.

Overall, this study showcases an effective approach to analyzing online content and discourse on nuclear energy-related topics using big data, providing insights into the opinions of an engaged online demographic. While this online demographic may not fully represent the Japanese public, these findings can complement traditional survey data to offer a more nuanced understanding of public viewpoints and inform efforts toward the continued development of domestic nuclear energy.

\section*{Acknowledgment}
A part of this work was supported by a research grant from Kangenkon. 

\section*{Competing Interests} The authors have no competing interests to declare.

\section*{Data Availability} The datasets and code used to generated the results in this study are available from Y. Sun upon reasonable request.


\begin{thebibliography}{10}
\providecommand{\url}[1]{\normalfont{#1}}
\providecommand{\urlprefix}{Available from: }

\bibitem{ANRE2021}
{Agency for Natural Resources and Energy}. \begin{CJK}{UTF8}{min}国内エネルギー動向\end{CJK} ; 2021. Accessed: 2024-11-14; \urlprefix\url{https://www.enecho.meti.go.jp/about/whitepaper/2021/pdf/2_1.pdf}.

\bibitem{METI2022}
{Agency for Natural Resources and Energy}. \begin{CJK}{UTF8}{min}今後の原子力政策について\end{CJK} ; 2022. Accessed: 2024-11-14; \urlprefix\url{https://www.meti.go.jp/shingikai/enecho/denryoku_gas/genshiryoku/pdf/024_03_00.pdf}.

\bibitem{nhk2022}
NHK. {\begin{CJK}{UTF8}{min}原発運転期間延長などの指針 賛成45\% 反対37\% NHK世論調査\end{CJK}} ; 2022. Accessed: 2024-11-14; \urlprefix\url{https://www3.nhk.or.jp/news/html/20221213/k10013921151000.html}.

\bibitem{asahi2024}
{The Asahi Shimbun}. \begin{CJK}{UTF8}{min}原発再稼働賛成50\% 反対35\%を上回る 朝日世論調査\end{CJK} ; 2024. Accessed: 2024-11-14; \urlprefix\url{https://www.asahi.com/articles/ASS2L7SJRS2HUZPS006.html}.

\bibitem{nikkei2023}
{The Nikkei}. \begin{CJK}{UTF8}{min}福島第一原発の処理水放出、賛成58\%\end{CJK} ; 2023. Accessed: 2024-11-14; \urlprefix\url{https://www.nikkei.com/article/DGXZQOUA284690Y3A720C2000000/}.

\bibitem{kitada2016public}
Kitada~A. {Public opinion changes after the Fukushima Daiichi Nuclear Power Plant accident to nuclear power generation as seen in continuous polls over the past 30 years}. J Nucl Sci Technol. 2016;\hspace{0pt}53(11):1686--1700.

\bibitem{ikegami2013topic}
Ikegami~Y, Kawai~K, Namihira~Y, et~al. {Topic and Opinion Classification Based Information Credibility Analysis on Twitter}. In: 2013 IEEE International Conference on Systems, Man, and Cybernetics; 2013. p. 4676--4681.

\bibitem{kim2014public}
Kim~D, Kim~JW. {Public opinion mining on social media: A case study of Twitter opinion on nuclear power}. Adv Sci Technol Lett. 2014;\hspace{0pt}51:224--228.

\bibitem{park2019positive}
Park~E. {Positive or negative? Public perceptions of nuclear energy in South Korea: Evidence from Big Data}. Nucl Eng Technol. 2019;\hspace{0pt}51(2):626--630.

\bibitem{jeong2021sentiment}
Jeong~SY, Kim~JW, Kim~YS, et~al. {Sentiment analysis of nuclear energy-related articles and their comments on a portal site in Rep. of Korea in 2010--2019}. Nucl Eng Technol. 2021;\hspace{0pt}53(3):1013--1019.

\bibitem{pu2022chinese}
Pu~X, Jiang~Q, Fan~B. {Chinese public opinion on Japan's nuclear wastewater discharge: A case study of Weibo comments based on a thematic model}. Ocean Coast Manag. 2022;\hspace{0pt}225. 106188.

\bibitem{hasegawa2020changing}
Hasegawa~S, Suzuki~T, Yagahara~A, et~al. {Changing Emotions About Fukushima Related to the Fukushima Nuclear Power Station Accident-How Rumors Determined People's Attitudes: Social Media Sentiment Analysis}. J Med Internet Res. 2020;\hspace{0pt}22(9):e18662.

\bibitem{kwon2024sentiment}
Kwon~OH, Vu~K, Bhargava~N, et~al. {Sentiment analysis of the United States public support of nuclear power on social media using large language models}. Renew Sustain Energy Rev. 2024;\hspace{0pt}200. 114570.

\bibitem{youtube-transcript-api}
Depoix~J. Youtube transcript api ; 2023. Accessed: 2024-11-25; \urlprefix\url{https://github.com/jdepoix/youtube-transcript-api}.

\bibitem{luca_papariello_2024}
{Luca Papariello}. xlm-roberta-base-language-detection ; 2024. Accessed: 2024-11-14; \urlprefix\url{https://huggingface.co/papluca/xlm-roberta-base-language-detection}.

\bibitem{martin2015more}
Martin~F, Johnson~M. {More Efficient Topic Modelling Through a Noun Only Approach}. In: Proceedings of the Australasian Language Technology Association Workshop 2015; 2015. p. 111--115.

\bibitem{may2019analysislemmatizationtopicmodels}
May~C, Cotterell~R, Durme~BV. {An Analysis of Lemmatization on Topic Models of Morphologically Rich Language} ; 2019. \urlprefix\url{https://arxiv.org/abs/1608.03995}.

\bibitem{sudachi}
Takaoka~K, Hisamoto~S, Kawahara~N, et~al. {Sudachi: a Japanese Tokenizer for Business}. In: Proceedings of the Eleventh International Conference on Language Resources and Evaluation (LREC 2018); 2018.

\bibitem{blei2003latent}
Blei~DM, Ng~AY, Jordan~MI. {Latent Dirichlet Allocation}. J Mach Learn Res. 2003;\hspace{0pt}3(Jan):993--1022.

\bibitem{roder2015exploring}
R{\"o}der~M, Both~A, Hinneburg~A. {Exploring the Space of Topic Coherence Measures}. In: {Proceedings of the Eighth ACM International Conference on Web Search and Data Mining}; 2015. p. 399--408.

\bibitem{oseti}
Yukino~I. oseti ; 2023. Accessed: 2024-11-25; \urlprefix\url{https://github.com/ikegami-yukino/oseti}.

\bibitem{dict1}
Kobayashi~N, Inui~K, Matsumoto~Y, et~al. {Collecting Evaluative Expressions for Opinion Extraction}. J Nat Lang Process. 2005;\hspace{0pt}12(3):203--222.

\bibitem{dict2}
Higashiyama~M, Inui~K, Matsumoto~Y. {Learning Sentiment of Nouns from Selectional Preferences of Verbs and Adjectives,}. In: Proceedings of the 14th Annual Meeting of the Association for Natural Language Processing; 2008. p. 584--587.

\bibitem{vaswani2023attentionneed}
Vaswani~A, Shazeer~N, Parmar~N, et~al. {Attention Is All You Need} ; 2023. \urlprefix\url{https://arxiv.org/abs/1706.03762}.

\bibitem{koheiduck-bert}
Takami~K. bert-japanese-finetuned-sentiment ; 2022. Accessed: 2024-11-24; \urlprefix\url{https://huggingface.co/koheiduck/bert-japanese-finetuned-sentiment}.

\bibitem{christian-phu-bert}
Phu~C. bert-finetuned-japanese-sentiment ; 2023. Accessed: 2024-11-24; \urlprefix\url{https://huggingface.co/christian-phu/bert-finetuned-japanese-sentiment}.

\bibitem{brown2020languagemodelsfewshotlearners}
Brown~TB, Mann~B, Ryder~N, et~al. {Language Models are Few-Shot Learners} ; 2020. \urlprefix\url{https://arxiv.org/abs/2005.14165}.

\bibitem{zhang2023sentimentanalysiseralarge}
Zhang~W, Deng~Y, Liu~B, et~al. {Sentiment Analysis in the Era of Large Language Models: A Reality Check} ; 2023. \urlprefix\url{https://arxiv.org/abs/2305.15005}.

\bibitem{krugmann2024sentiment}
Krugmann~JO, Hartmann~J. {Sentiment Analysis in the Age of Generative AI}. Cust Needs Solut. 2024;\hspace{0pt}11. 3.

\bibitem{kheiri2023sentimentgptexploitinggptadvanced}
Kheiri~K, Karimi~H. {SentimentGPT: Exploiting GPT for Advanced Sentiment Analysis and its Departure from Current Machine Learning} ; 2023. \urlprefix\url{https://arxiv.org/abs/2307.10234}.

\bibitem{zhong2023chatgptunderstandtoocomparative}
Zhong~Q, Ding~L, Liu~J, et~al. {Can ChatGPT Understand Too? A Comparative Study on ChatGPT and Fine-tuned BERT} ; 2023. \urlprefix\url{https://arxiv.org/abs/2302.10198}.

\bibitem{zarrabeitia2023nuclear}
Zarrabeitia-Bilbao~E, Jaca-Madariaga~M, Rio-Belver~RM, et~al. {Nuclear energy: Twitter data mining for social listening analysis}. Soc Netw Anal Min. 2023;\hspace{0pt}13. 29.

\bibitem{jaero2022}
{Japan Atomic Energy Relations Organization}. \begin{CJK}{UTF8}{min}原子力に関する世論調査 （2022 年度）調査結果\end{CJK} ; 2022. Accessed: 2024-11-13; \urlprefix\url{https://www.jaero.or.jp/_files/poll/tyousakenkyu2022/results_2022.pdf}.

\bibitem{yagahara2018relationships}
Yagahara~A, Hanai~K, Hasegawa~S, et~al. {Relationships Among Tweets Related to Radiation: Visualization Using Co-Occurring Networks}. JMIR Public Health Surveill. 2018;\hspace{0pt}4(1):e26.

\bibitem{mitchell2013twitter}
Mitchell~A, Hitlin~P. {Twitter Reaction to Events Often at Odds with Overall Public Opinion} ; 2013. Accessed: 2024-11-25; \urlprefix\url{https://www.pewresearch.org/2013/03/04/twitter-reaction-to-events-often-at-odds-with-overall-public-opinion/}.

\bibitem{li2016tweeting}
Li~N, Akin~H, Su~LYF, et~al. {Tweeting disaster: an analysis of online discourse about nuclear power in the wake of the Fukushima Daiichi nuclear accident}. J Sci Commun. 2016;\hspace{0pt}15(5). A02.

\end{thebibliography}

\end{document}